\documentclass[letterpaper, 10pt, conference]{ieeeconf}  

\IEEEoverridecommandlockouts                              
\usepackage{amsmath} 
\usepackage{amssymb}  
\usepackage[utf8]{inputenc}
\usepackage[keeplastbox]{flushend}
\usepackage{graphicx}
\usepackage{cite}
\usepackage{siunitx}
\usepackage[caption=false]{subfig}
\usepackage{afterpage}
\usepackage{booktabs}
\usepackage{siunitx}
\usepackage{multicol}
\usepackage[bookmarks=true]{hyperref}
\usepackage{mathdots} 

\title{3D3L: Deep Learned 3D Keypoint Detection and Description for LiDARs
\author{Dominic Streiff$^1$, Lukas Bernreiter$^1$, Florian Tschopp$^1$, Marius Fehr$^{1,2}$ and Roland Siegwart$^1$} 
\thanks{This work was partially supported by the National Center of Competence in Research (NCCR) Robotics through the Swiss National Science Foundation and Siemens Mobility GmbH, Germany.}
\thanks{$^1$Autonomous Systems Lab, ETH Zurich, Zurich, Switzerland, {\tt \small \url{{dostreif,berlukas,ftschopp,rsiegwart}@ethz.ch}}}
\thanks{$^2$Voliro Airborne Robotics, Zurich, Switzerland, {\tt \small \url{marius.fehr@voliro.com}}} 
\thanks{For the benefit of the community, the code is available at \url{https://github.com/ethz-asl/3d3l}}
\thanks{\textcopyright 2021 IEEE. Personal use of this material is permitted. Permission from IEEE must be obtained for all other uses, in any current or future media, including reprinting/republishing this material for advertising or promotional purposes, creating new collective works, for resale or redistribution to servers or lists, or reuse of any copyrighted component of this work in other works
}
}

\usepackage[utf8]{inputenc}
\usepackage[OT1]{fontenc}

\usepackage{fancyhdr}

\usepackage{float}
\usepackage{textcomp}\usepackage{gensymb}

\usepackage{multicol}

\usepackage{subfig}
\usepackage{graphicx}

\usepackage{amsmath}
\usepackage{amssymb}

\usepackage{upgreek}

\usepackage{units}

\usepackage{pdfpages}
\includepdfset{pages={-}, frame=true, pagecommand={\thispagestyle{fancy}}}

\usepackage[nolist,nohyperlinks]{acronym}

\newcommand{\norm}[1]{\left\lVert#1\right\rVert}

\begin{acronym}
\acro{SLAM}{Simultaneous Localization And Mapping}
\acro{TSDF}{Truncated Signed Distance Field}
\acro{ICP}{Iterative Closest Point}
\acro{RFS}{Random Finite Sets}
\acro{CNN}{Convolutional Neural Network}
\acro{CDF}{Cumulative Distribution Function }
\acro{FCGF}{fully convolutional geometric feature}
\acro{NMS}{non-maxima supression}
\acro{ASL}{Autonomous Systems Lab}
\acro{BoVW}{bag of visual words}
\end{acronym}

\begin{document}

\maketitle

\begin{abstract}
With the advent of powerful, light-weight 3D LiDARs, they have become the hearth of many navigation and SLAM algorithms on various autonomous systems.
Pointcloud registration methods working with unstructured pointclouds such as ICP are often computationally expensive or require a good initial guess. 
Furthermore, 3D feature-based registration methods have never quite reached the robustness of 2D methods in visual SLAM.
With the continuously increasing resolution of LiDAR range images, these 2D methods not only become applicable but should exploit the illumination-independent modalities that come with it, such as depth and intensity.
In visual SLAM, deep learned 2D features and descriptors perform exceptionally well compared to traditional methods. In this publication, we use a state-of-the-art 2D feature network as a basis for 3D3L, exploiting both intensity and depth of LiDAR range images to extract powerful 3D features.
Our results show that these keypoints and descriptors extracted from LiDAR scan images outperform state-of-the-art on different benchmark metrics and allow for robust scan-to-scan alignment as well as global localization.
\end{abstract}

%
%
\section{Introduction}
\label{sec:introduction}
Pointcloud registration aims at finding a rigid transformation between two partially overlapping pointclouds and constitutes a critical task in pointcloud-based \ac{SLAM}. 
It is particularly crucial for ill-lighted and unstructured environments where illumination invariance is vital for a successful mapping and localization system~\cite{Dong2014}.
Traditional methods such as \ac{ICP}~\cite{Rusinkiewicz2001} operate on unordered pointclouds and iteratively refine the point correspondences and the rigid transformation at each step. 
Consequently, these solutions degrade in the presence of high noise and motion distortions due to local minima in the optimization and thus need a good prior to succeed.
In contrast, feature-based registration systems directly compute interest points and descriptors to match corresponding input data points between scans. 
One of the advantages of these feature-based methods are a sparse representation of information needed, i.e. for global loop closure. Moreover, it is possible to leverage the matured research in 2D feature-based methods.
Current visual localization and place recognition systems have shown great improvement by jointly learning the keypoint detector and descriptor~\cite{Dusmanu2019, Detone2018}. 
Especially, since keypoints in non-salient regions are easily mismatched or poorly localized which results in bad transformation estimation results. 
However, most LiDAR-based counterparts emphasize learning local descriptors only and randomly sample keypoints from the pointcloud~\cite{Choy2019}.
%
\begin{figure}[!t]
    \centering
    \includegraphics[width=\columnwidth]{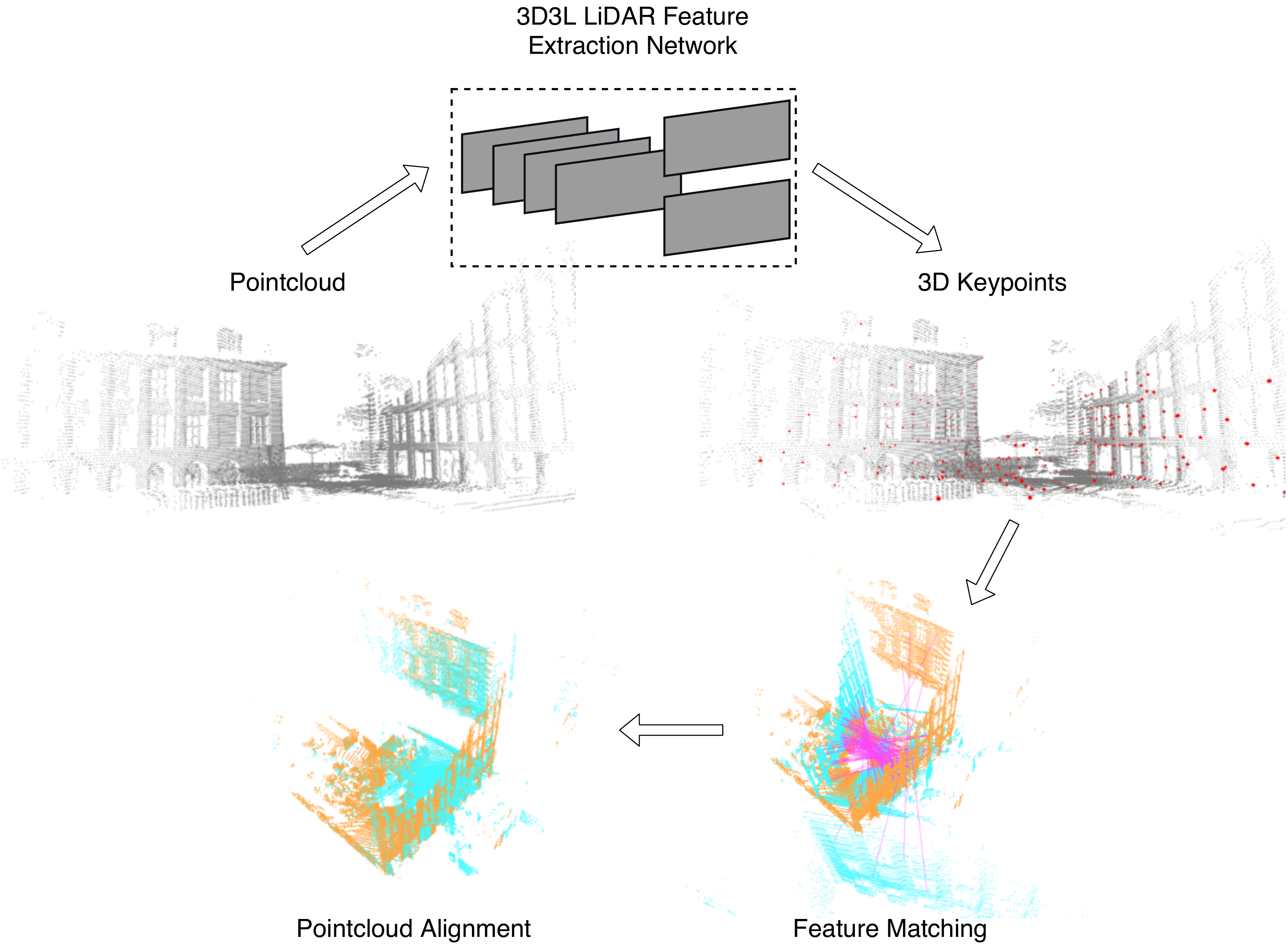}
    \vspace{-5mm}
    \caption{We propose a jointly learned 3D LiDAR keypoint detection and description pipeline suitable for the task of global pointcloud alignment and global localization.}
    \label{fig:range_inte_intro}
    \vspace{-7mm}
\end{figure}
Inspired by deep learned RGB image feature extraction methods, we present 3D3L, a network which jointly detects and describes keypoints on 3D LiDAR scans.
Instead of using unordered pointclouds to extract features from, we project and combine the range and intensity channels from LiDAR scans to images.
The advantage of using scan images is that we use the information of the LiDAR scan's ordering. Moreover, we can easily  use intensity information besides range information only. The channels of an RGB image represent inherently different modalities than the range and intensity channels and thus need to be processed differently. Therefore, we need to adapt the RGB feature extraction network and the training procedure to extract 3D features from a scan image.

Our contributions are as follows:
\begin{itemize}
    \item Adaptation of a state-of-the-art image feature network to use-case of multi-modal LiDAR scan images.
    \item A training procedure using LiDAR scan images with ground-truth pixel correspondences to produce reliable and repeatable 3D features.
    This leads to a pipeline that is outperforming state-of-the-art keypoint/descriptor methods in terms of repeatability and matching/registration recall on LiDAR data.
    \item Demonstrating the versatility of these features by applying them in a proof-of-concept LiDAR \ac{SLAM} system for both scan-to-scan pose estimation as well as loop closure.
\end{itemize}

\section{Related Work}
\label{sec:related_work}
Pointcloud registration is a well-studied field for various applications. An extensive overview is given by H{\"{a}}nsch et al.~\cite{Hansch2014} in the context of feature-based alignment.
In this section, we will focus on methods related to the detection and description of interest points in pointclouds and LiDAR scans.
We structure it into interest point detectors, local and global approaches for description, and conclude with methods that perform both jointly.
\subsection{Interest Point Detectors}
\label{sec:detectors}
Even today many of the interest point detection approaches rely on simple, hand-crafted methods, such as Harris3D~\cite{Sipiran2011}, ISS~\cite{Zhong2009} or even uniform random sampling.
Several advances in the 2D image domain have shown that data-driven methods have the potential to significantly outperform hand-crafted solutions~\cite{Dusmanu2019}.
The Unsupervised Stable Interest Point~\cite{Li2019} is an unsupervised learning-based 3D keypoint detector. This approach shows that learning 3D interest points can also outperform known hand-crafted methods.
\subsection{Descriptors}
\label{sec:descriptors}
Local pointcloud descriptors use only a neighborhood around a given interest point to describe it, whereas global descriptors exploit the whole pointcloud. 
Traditional local approaches like SHOT~\cite{Salti2014} and NARF~\cite{Steder2010} use hand-crafted description techniques such as point histograms. 
However, recent advances in data-driven methods for describing local descriptors have shown to perform significantly better than hand-crafted approaches. 
Zeng et al.~\cite{Zeng2017} proposed a neural network which describes local 3D patches around given interest point, converting them into a voxel grid of truncated distance function values. 
Deng et al.~\cite{Deng2019} extract local descriptors from unordered pointcloud patches and directly train to minimize a local (patch-wise) pose estimation error. 
Both methods solely rely on local patch information around the interest points, which reduces the network's receptive field.
Furthermore, the performance of these methods is dependent the quality of the external source of interest points.\\
Choy et al.~\cite{Choy2019} present a \ac{FCGF} network for dense feature description on pointclouds. They introduce a custom 3D convolution to extract dense descriptors for the whole pointcloud, while relying on random sampling for keypoint extraction.
\subsection{Joint Detection and Description}
\label{sec:learned_detection_description}
In the 2D image domain it was shown that jointly detecting and describing interest points can improve the descriptors' quality~\cite{Dusmanu2019}.
Similarly for pointclouds, 3DFeat-Net~\cite{Yew2018} and D3Feat~\cite{Bai2020} propose a pipeline to jointly train a 3D feature extraction and description. 
3DFeat-Net~\cite{Yew2018} is a two-stage network which first calculates keypoints and then the descriptors for the selected keypoints only. 
Whereas, D3Feat~\cite{Bai2020} computes a dense map for interest points and descriptors simultaneously using the custom 3D convolution presented by \ac{FCGF} \cite{Choy2019}.

Both 3dFeat-Net and D3Feat train on pointclouds directly, whereas to our knowledge, nobody has yet published work that focuses on jointly learning keypoint extraction and description on the combined range and intensity images from LiDAR scanners. 
\section{Method}
\label{sec:method}
In this section, we discuss the main components of our proposed LiDAR-based keypoint detector and descriptor framework. 
Initially, we will give an overview of the detailed network structure and pipeline used for generating suitable pixel correspondences between pairs of scan images.
Furthermore, we will outline details on our training procedure and the methodology to use our pipeline for pose estimation.
\subsection{Feature Extraction Network}
\label{chp:feat_extr}
Inspired by recent advances in data-driven approaches for RGB images, we first discuss methodologies to extract features from a range and intensity image.
SuperPoint~\cite{Detone2018} has a semi-supervised pre-training procedure for the keypoint locations, for which an adaption to the LiDAR case is challenging.
Similarly, UnSuperPoint~\cite{Christiansen2019} uses the random homography of the training pairs as matching information in the loss function, which can not be easily adapted to the LiDAR scan domain.
In contrast, the R2D2 network~\cite{Revaud2019} uses a pixel matching map for ground truth correspondences during training, which can readily be adapted to work with LiDAR scan images and also achieves state-of-the-art performance~\cite{Revaud2019}.
\subsubsection{Network Structure}
\label{sec:net_strucuture}
\begin{figure}
    \vspace{2mm}
    \centering
    \includegraphics[width=\columnwidth]{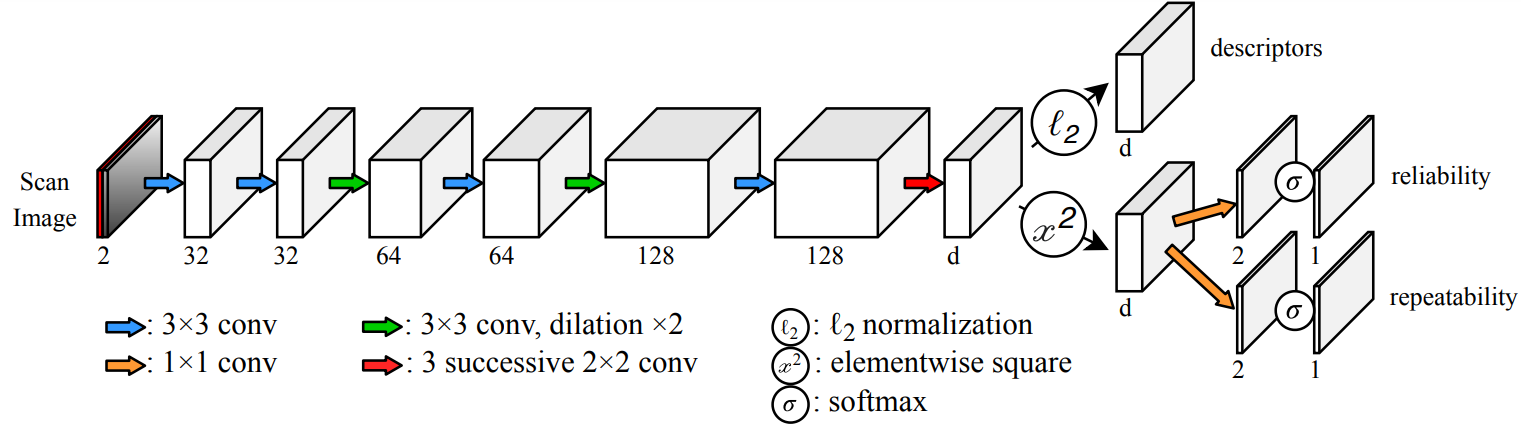}
    \vspace{-5mm}
    \caption{Overview of the network architecture adapted from the R2D2~\cite{Revaud2019} network.}
    \vspace{-2mm}
    \label{fig:net_structure}
\end{figure}
\begin{figure}
    \centering
    \includegraphics[width=\columnwidth]{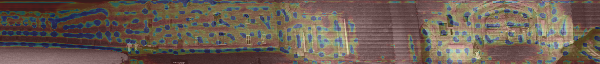}
    \vspace{-5mm}
    \caption{Intensity channel of LiDAR scan image with overlayed score map of the combined reliability and repeatability outputs.}
    \vspace{-6mm}
    \label{fig:scoremap}
\end{figure}
The R2D2 network comprises a fully convolutional structure optimized to extract 128-dimensional descriptors from RGB images along with score maps based on the reliability and repeatability scores. Reliability describes how distinct the key point is based on its local neighbourhood. Repeatability is the measure of how precise a keypoint is located compared to the same keypoint from another viewpoint.
We modified the network structure (cf. Figure~\ref{fig:net_structure}) to take range and intensity images from LiDAR scans as input and decreased to a 32-dimensional descriptor for fast lookup.
\begin{figure}
    \vspace{2mm}
    \centering
    \includegraphics[width=\columnwidth]{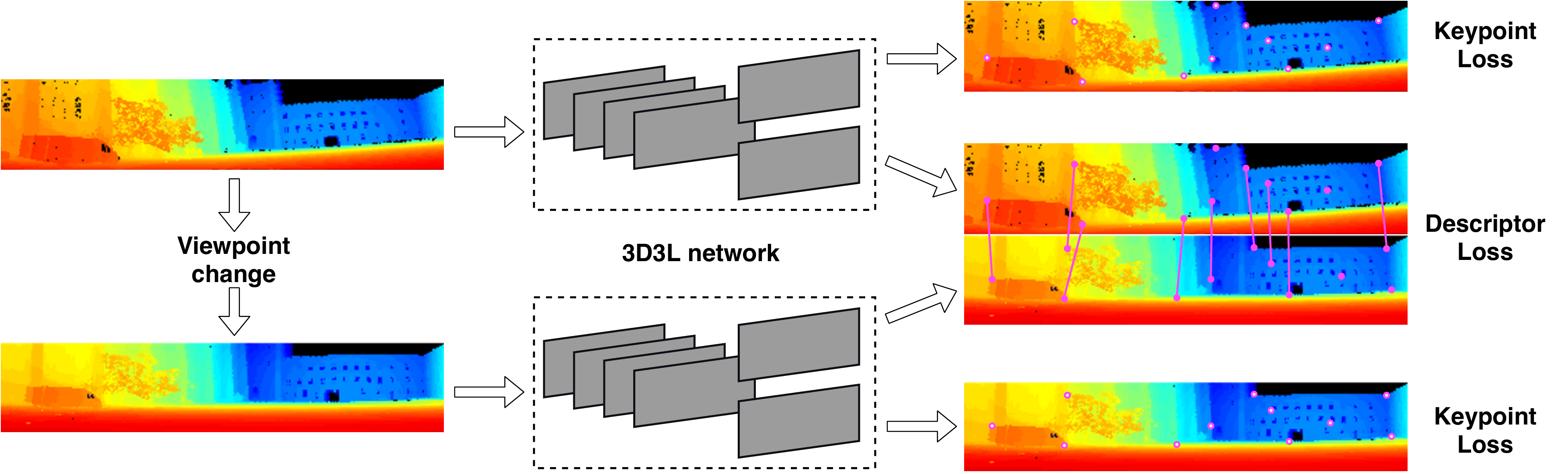}
    \vspace{-5mm}
    \caption{Illustration of the twin network set-up used for training our network.}
    \label{fig:twin_net}
    \vspace{-6mm}
\end{figure}
We utilized a twin network approach for training our 3D3L network as proposed in~\cite{Detone2018}. 
Specifically, a pair of different input images of the same scene $I$ and $I'$ are processed by the network simultaneously as seen in Figure~\ref{fig:twin_net}. 
We have not modified the loss function and refer the interested reader to the original R2D2 paper~\cite{Revaud2019}.
To calculate the loss,  our network requires a ground-truth correspondence map $\mathbf{U} \in \mathbb{R}^{H\times W\times 2}$.
Here, $\mathbf{U}_{u,v}$ maps the pixel $(u,v)$ in the first image $I$ to the corresponding location $(u',v')$ in image $I'$.
\subsubsection{Output Processing}
\label{sec:out_processing}
The outcome of the network needs to be processed from dense scores and descriptors in the scan image to actual 3D keypoints and their corresponding descriptors.
First, the repeatability and reliability score maps (Figure~\ref{fig:scoremap} are multiplied to have a single keypoint score map $\mathbf{S} \in [0, 1]^{H \times W}$ and pixel locations with a score value larger than the threshold $\tau_{score}$ are selected.  
Then, \ac{NMS} with a radius $N_{nms}$ is applied. For each selected pixel $(u,v)$ of the score map $\mathbf{S}_{u,v}$ the corresponding descriptors $\mathbf{X}_{u,v}$ are selected. Since the LiDAR scan offers an ordered pointcloud $\mathbf{P} \in \mathbb{R}^{H\times W \times 3}$ the corresponding 3D location of the selected pixel $\mathbf{P}_{u,v}$ can be easily extracted. Not every point $\mathbf{P}_{u,v}$ lies in a valid region of the LiDAR scan (non-returned or deflected LiDAR rays). Therefore the invalid keypoints $\mathbf{P}_{u',v'}$ are removed from the list.
This final selection of 3D keypoints $\mathbf{P}_{u,v}$, and the corresponding descriptors $\mathbf{X}_{u,v}$ are passed on to be further processed.
\subsection{Scan Images}
\label{sec:scan_images}
We use both the range and intensity images from a LiDAR scan as input for the the feature extraction network.
This enables to leverage both geometrical and texture information of the environment. The LiDAR scanner offers an ordered pointcloud $\mathbf{P} \in \mathbb{R}^{H \times W \times 3}$ and corresponding intensities $L \in \mathbb{R}^{H \times W}$. 
The pointcloud can be transformed into a range image $R \in \mathbb{R}^{H \times W}$, such that $R_{i,j} = \norm{\mathbf{P}_{i,j}}$. \\
We call the 2-channel image consisting of the range image $R$ and intensity image $L$, \textit{scan image} $I$, which is the input image of the network.\\

Generally, we require a pair of scan images $I$ and $I'$ showing the same scene from different viewpoints for training. 
Moreover, the ground truth pixel match $\mathbf{U} \in \mathbb{R}^{H \times W \times2}$ is required to calculate the loss. The pixel match describes the location of each pixel $(u,v)$ in image $I$ in the other image $I'$. \\
We use two different approaches to generate a training data $(I, I', \mathbf{U})$, synthetic and real scan pairs.
\\
\subsubsection{Synthetic Scan Pairs}
\label{subsec:synthetic_pairs}
The idea to use synthetically generated training pairs is inspired by the RGB case. There, random affine projections and illumination changes are applied to generate the training data from just a single image. We adapted this approach to LiDAR scan images. Using the same point cloud and project it to a scan frame in another viewpoint lead to many undefined regions due to missing values, resulting in non-converging training.

The following image transformations are applied to the scan image:
\begin{itemize}
\item \textit{Scale:} Scaling the scan image will affect the range and intensity channels differently. The intensity channel can be viewed as a grayscale image, therefore scaling the image can be solved by interpolating missing values. However, the range image needs to be altered additionally. Zooming in, for example, means being closer to an object, thus the range value will be smaller than before. We use an approximation of this transformation by multiplying with the scaling factor $s$. The scaled range image $R_s$ is calculated from the original range image $R$, such that $R_s = \frac{R}{s}$.
\item \textit{Translation}: The translation must be split in $u-$ and $v-$translation. Whereas translating the image in $v-$direction is limited by the height of the image, since a reasonable amount of the image should be inside the image boundary of image $I'$, $u-$translation is unlimited since LiDAR scan images are a 360° view, meaning the right edge of the image can be seamlessly continued with the left edges pixels. Therefore, shifting in $u-$direction means rotating the LiDAR scanner around the $z-$axis. The pixels that are shifted outside the image boundary are wrapped to the other edge.
\item \textit{Tilting:} Tilting the image also means a viewpoint change of the LiDAR camera. Therefore, the range values change as well. However, the model for shearing is even more complex than only zooming, thus we keep the shearing range small to avoid these problems and don't change the range values.
\end{itemize}
With the known affine transformation, the pixel match $\mathbf{U}$ can be directly calculated. Pixels that are transformed outside the image boundaries will be marked as invalid and are ignored by the loss function.
\\
\subsubsection{Real Scan Pairs}
\label{subsec:real_pairs}
\begin{figure}
    \vspace{2mm}
    \begin{minipage}{\linewidth}
    \includegraphics[width=\linewidth]{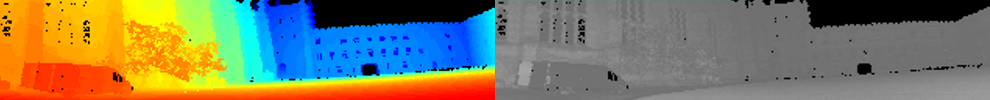}
    \vspace{-3mm}
    \end{minipage}
    \begin{minipage}{\linewidth}%
    \includegraphics[width=\linewidth]{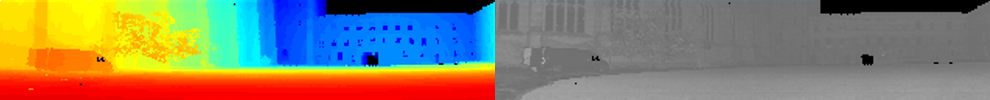}
    \vspace{-3mm}
    \end{minipage}
    \begin{minipage}{\linewidth}%
    \includegraphics[width=\linewidth]{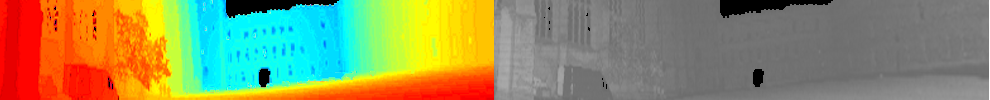}
    \end{minipage}
    \vspace{-3mm}
    \caption{Range and intensity channels of scan images $I$ (top) and $I'$ for the real (middle) and synthetic (bottom) pairs on the \textit{Newer College Dataset}.}
    \label{fig:real_pairs_range}
    \vspace{-6mm}
\end{figure}

The second method makes use of the known transformation between two different scans of the same scene from a dataset with ground truth poses. An example of two scan images $I$ and $I'$ that show the same scene is seen in Figure~\ref{fig:real_pairs_range}. 

\textbf{Selecting pairs:} First, suitable pairs of scan images need to be found. Knowing the ground truth pose of the LiDAR scanner for each scan image enables a proximity-based approach. We select the first image $I$ recorded at the pose $\mathbf{p_L} \in \mathbb{R}^{4 \times 4}$. Then, a second image recorded at the pose $\mathbf{p_L'} \in \mathbb{R}^{4 \times 4}$ is sampled from all poses that are inside the spherical shell with an inner radius $r_i$ and outer radius $r_o$ centered at $\mathbf{p_L}$. Pairs selected this way mostly show a similar scene, while still being different enough because of the minimal distance $r_i$ between them.\\
However, sometimes the scene is still not similar enough to successfully train the network. Therefore, we check how much the two pointclouds $\mathbf{P}$ and $\mathbf{P'}$ overlap. The overlap is calculated as follows: pointcloud $\mathbf{P}$ is transformed into the frame of pointcloud $\mathbf{P'}$, $\mathbf{P_T} = \mathbf{P}\mathbf{T_L}$, using the known transformation $\mathbf{T_L} = (\mathbf{p_L})^{-1}\mathbf{p_L'}$. Then, a nearest neighbour search for each point in pointcloud $\mathbf{P_T}$ in pointcloud $\mathbf{P'}$ is executed. If the nearest neighbour for a point is below a threshold, it is defined to have a corresponding point in the other frame. The overlap $\Omega$ is then calculated as follows $\Omega = \frac{N_i}{N_t}$, where $N_i$ is the number of points in pointcloud $\mathbf{P}$ that have a corresponding point in pointcloud $\mathbf{P'}$ and $N_t$ is the number of total points in pointcloud $\mathbf{P}$. If $\Omega > \tau_{o}$ the two images are accepted as a training pair, otherwise another scan $I'$ is sampled.
\\
\textbf{Pixel flow:} Besides the two scan images, the pixel flow $\mathbf{U}$ is required to train the network. We take the transformed pointcloud $\mathbf{P_T}$ and project it to a range image using spherical projection. This results in a pair of pixel coordinates $(u',v')$ for each pixel $(u,v)$ for every valid point in image $I$. The intensities from image $I$ are mapped to the scan image frame of image $I'$, using $\mathbf{U}$. Due to viewpoint changes multiple points $(u,v)$ may map to a single point $(u',v')$ or there are missing correspondences, resulting in non-existing pixel correspondences represented in black. Invalid points in the first pointcloud are not mapped at all (no 3D coordinates available to transform) and points mapped to invalid regions in image $I'$ are removed from the flow map.
Moreover, occlusion of points is handled as follows. If the range of corresponding pixel $(u', v')$ in image $I'$ is closer than a threshold from the corresponding range value of image $I$ the point is defined to be occluded and thus removed from the pixel flow. 
\\
After all those steps, the training data $(I, I', \mathbf{U})$ extracted only from real LiDAR scans can be used to train the network.
\subsection{Training Details}
\label{chp:training_details}
\subsubsection{Dataset}
\label{sec:dataset}
We extract the training data from several datasets comprising indoor and outdoor scenes that were recorded with state-of-the-art 3D LiDAR scanners, an Ouster OS-1 64 (65535 points per scan) and an Ouster OS-0 128 (131072 points per scan).
\\
\textbf{The Newer College Dataset} by Ramezani et al.~\cite{ramezani2020} is an outdoor dataset including data from a variety of mobile mapping sensors collected using a handheld device carried through New College, Oxford, UK. It has centimeter precise ground truth odometry data provided by a Leica BLK scanner. 
The employed LiDAR data is a 64 beam Ouster OS-1 scanner. Since the scans are available at 10 Hz, many scans show nearly the same scene. Therefore, we only use every tenth scan, resulting in approximately 2600 different scans.\\
\textbf{Self-recorded datasets}
We additionally utilize a series of self-recorded datasets comprising indoor and outdoor environments.
\begin{itemize}
  \item ASL indoor OS-1: This dataset was recorded in the \ac{ASL} at ETH Zurich with a 64 beam Ouster OS-1 scanner. It features hallways and some larger rooms and consists of approximately 400 scans.
  \item ETH terrace OS-0: This dataset was recorded inside and on a rooftop terrace at ETH Zurich with a 128 beam Ouster OS-0 scanner. It includes data of hallways, stairways and outdoor scenes of the large rooftop terrace and consists of approximately 500 scans.
  \item ETH outside OS-0: Recorded around some buildings of ETH Zurich with a 128 beam Ouster OS0 scanner, this part of the dataset mostly shows streets and buildings and consists of approximately 800 scans.
\end{itemize}
For all recordings, we created a global multi-session map using  Maplab~\cite{schneider2018} and optimized it using constraints from visual landmarks, IMU and LiDAR (\ac{ICP}) to serve as our ground truth. 
The ASL indoor OS-1 dataset is not used for training, only for evaluation (see chapter \ref{sec:experiments}). The data recorded with the 128 beam scanner contains 8 static regions with invalid data from a camera mount. In order to achieve stable training, the invalid sections were cropped. This cropping made the usage of those scans with the real scan pairs approach impossible, since the correspondences mostly lie outside these cropped regions.

Combining all datasets, we ended up with the following training setups to compare:
(1) A network trained with synthetic pairs comprising the newer college, ETH terrace and ETH outside datasets and (2) a network trained with only the newer college datasets real pair scans.

\subsubsection{Training Parameters}
\label{sec:parameters}
\begin{itemize}
    \item \textit{Preprocessing:}
    The LiDAR scan images are normalized to have zero mean and unit variance, where the mean and variance are evaluated per dataset. Moreover, only a randomly cropped region is used, which is selected based on the overlap of the images obtained from the pixel flow, to enable efficient training. The crop size is chosen to be $\unit[64 \times 180]{px}$. Random noise with zero mean and variance of 0.04 is added to both channels and both channels pixel values are offset independently by a random amount in the interval $[-0.1, 0.1]$ and scaled randomly with a factor in the interval $[-0.1, 0.1]$.\\
    The synthetic pairs are generated with following the parameters: random scaling factor in the interval $[1, 1.25]$, no $v-$translation and $u-$translation of $\leq \unit[50]{px}$, and tilting of $\leq \unit[20]{px}$.

    \item \textit{Network parameters:}
    The hyper-parameters of the network are mostly the same as in the original R2D2~\cite{Revaud2019} network. Only the patch size $N$ of the loss function is reduced to $\unit[8]{px}$, since we use inputs with lower resolution than possible with RGB images. Moreover, the batch size is reduced to $4$. We trained both networks (synthetic and real pairs) in two stages. First 3 epochs on only the synthetic pairs from the newer college dataset with no pixel value offset and scaling applied. Then with 20 epochs on the datasets described above. Directly training with large distortions resulted in the network to diverge during training.
    
    \item \textit{Post-processing:}
    To extract keypoints and descriptors from the dense outputs of the network as described in Section \ref{sec:out_processing}, the keypoint score threshold is $\tau_{score}=0.7$ and the NMS pixel radius is $N_{nms}=8$. These values are fixed based on empirical evaluation of keypoint distribution and quality.
\end{itemize}
\subsection{Pose Estimation}
\label{sec:pose_estimation}
\begin{figure}
    \vspace{2mm}
    \centering
    \includegraphics[width=\linewidth]{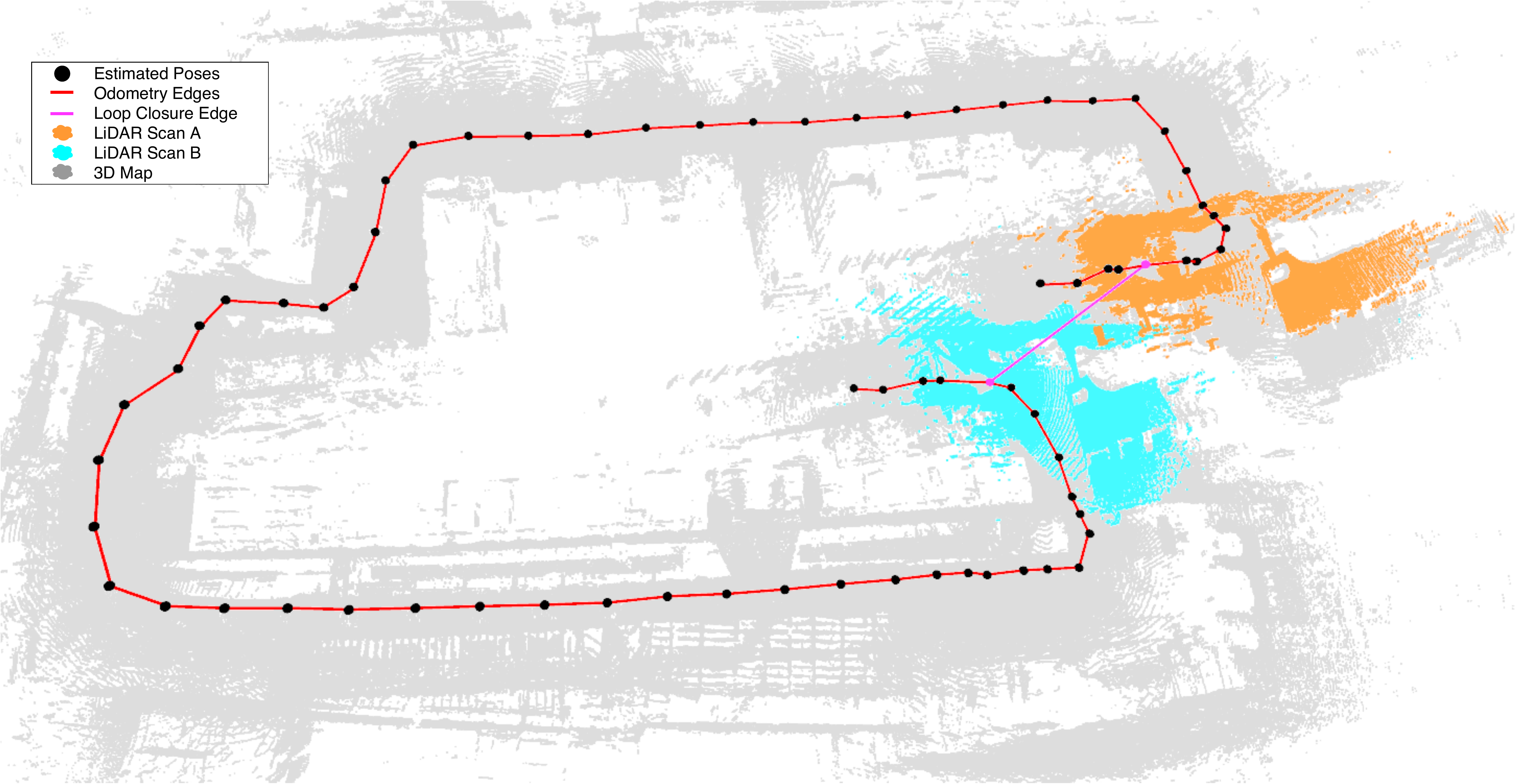}
    \vspace{-5mm}
    \caption{Pose graph of the estimated trajectory on the ETH outside dataset with pointclouds and exemplary loop closure edge.}
    \vspace{-5mm}
    \label{fig:pose_graph}
\end{figure}

The pose estimation of a series of following LiDAR scans is achieved by estimating the transformation between two following LiDAR scans. Thus, a simple pose graph can be built from an initial pose, and the scan-to-scan relative transformation estimates.\\
Moreover, the pose graph can be extended by loop closure edges, as seen in Figure \ref{fig:pose_graph}. The main idea is to find overlapping scenes with a \ac{BoVW} approach, then use the extracted features to estimate the transformation between the overlapping scenes and add it as a loop closure constraint to the pose graph.
\begin{itemize}
\item{\textit{\Acf{BoVW}:}}
For each scan image, the extracted descriptors are put in a list, on which k-means clustering in descriptor space is applied. The k clusters can be interpreted as visual words. Each descriptor is thus allocated a word (cluster). Counting the words per scan image leads to a histogram for every scan, which describes the image's visual content. The histograms are normalized, and then the Euclidean distance between each pair of histograms is calculated. If the distance is below a threshold $\tau_h$ the pair of scan images are considered to be overlapping. 
\item{\textit{Loop closure edges:}}
Each pair of overlapping scans is added as a loop closure edge to the pose graph. The loop closure constraint is the estimated rigid transformation $T_{LC}$ between the two pointclouds.
\item{\textit{Pose graph optimization:}}
The final pose graph with odometry and loop closure constraints is optimized with the Levenberg-Marquardt algorithm~\cite{Levenberg1944}. We use the implementation of Open3D~\cite{Zhou2018}.
\end{itemize}
\section{Experiments}
\label{sec:experiments}
This section evaluates the accuracy and effectiveness of our proposed keypoint detection and description pipeline. 
First, we validate the precision and repeatability of our keypoint detection using well-known benchmarks. 
Finally, we will integrate our pipeline into a SLAM framework where we use it to estimate the odometry and as a loop closure engine using a \ac{BoVW} approach and evaluate it using \cite{Zhang2018}.
\subsection{Evaluation Metrics}
Generally, we employ three different benchmark metrics to evaluate our network~\cite{Choy2019}: (i) The \textit{repeatability score}, i.e., the percentage of inlier keypoints which are defined as the keypoints with a maximum distance of $\tau_1=\unit[0.3]{m}$ after transforming with the ground truth alignment transformation $\mathbf{T_{gt}}$. (ii) The matching score, i.e., the fraction of scan pairs, which have more than $\tau_2=0.2$ correctly estimated matches. (iii) The \textit{registration recall} which is the percentage of scan pairs, for which the translation error is below the threshold $\tau_3=0.3$. 
\subsection{Repeatability and Reliability}
\label{sec:repeatability_reliability_results}
\subsubsection{Evaluation Dataset}
 We have used the ASL indoor OS1 dataset for evaluation and generated pairs as explained in Section~\ref{subsec:real_pairs} with an overlap threshold of $\tau_o=0.2$. To have more variety in the images, only every 10-th frame ($\unit[1]{Hz}$) was used as an anchor to find a corresponding scan. This resulted in 127 evaluation pairs $(I, I')$.
\subsubsection{Results}
\begin{figure}
    \vspace{2mm}
    \begin{minipage}{0.49\linewidth}
    \includegraphics[width=\linewidth]{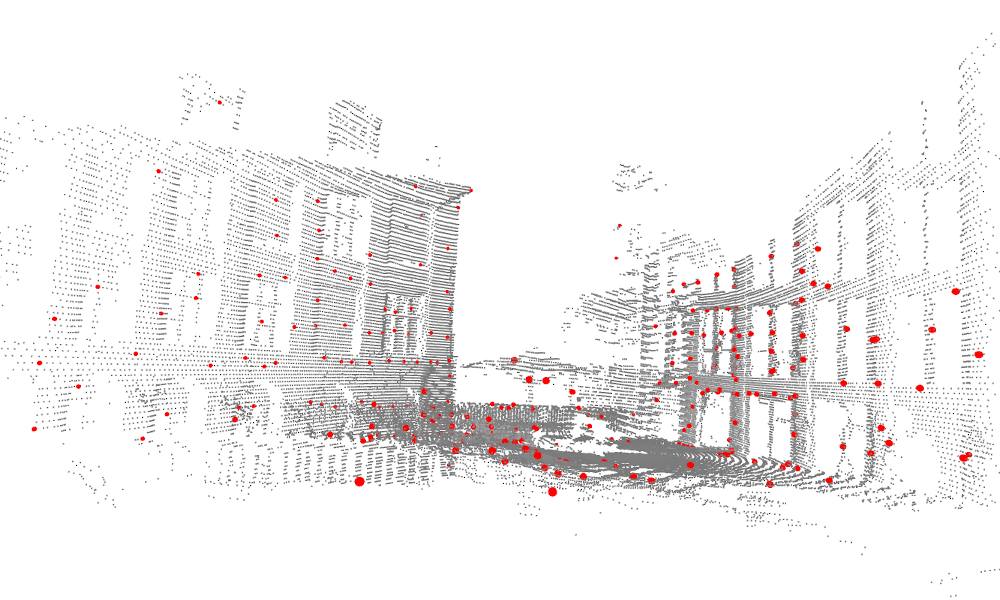}
    \end{minipage}\hfill
    \begin{minipage}{0.49\linewidth}%
    \includegraphics[width=\linewidth]{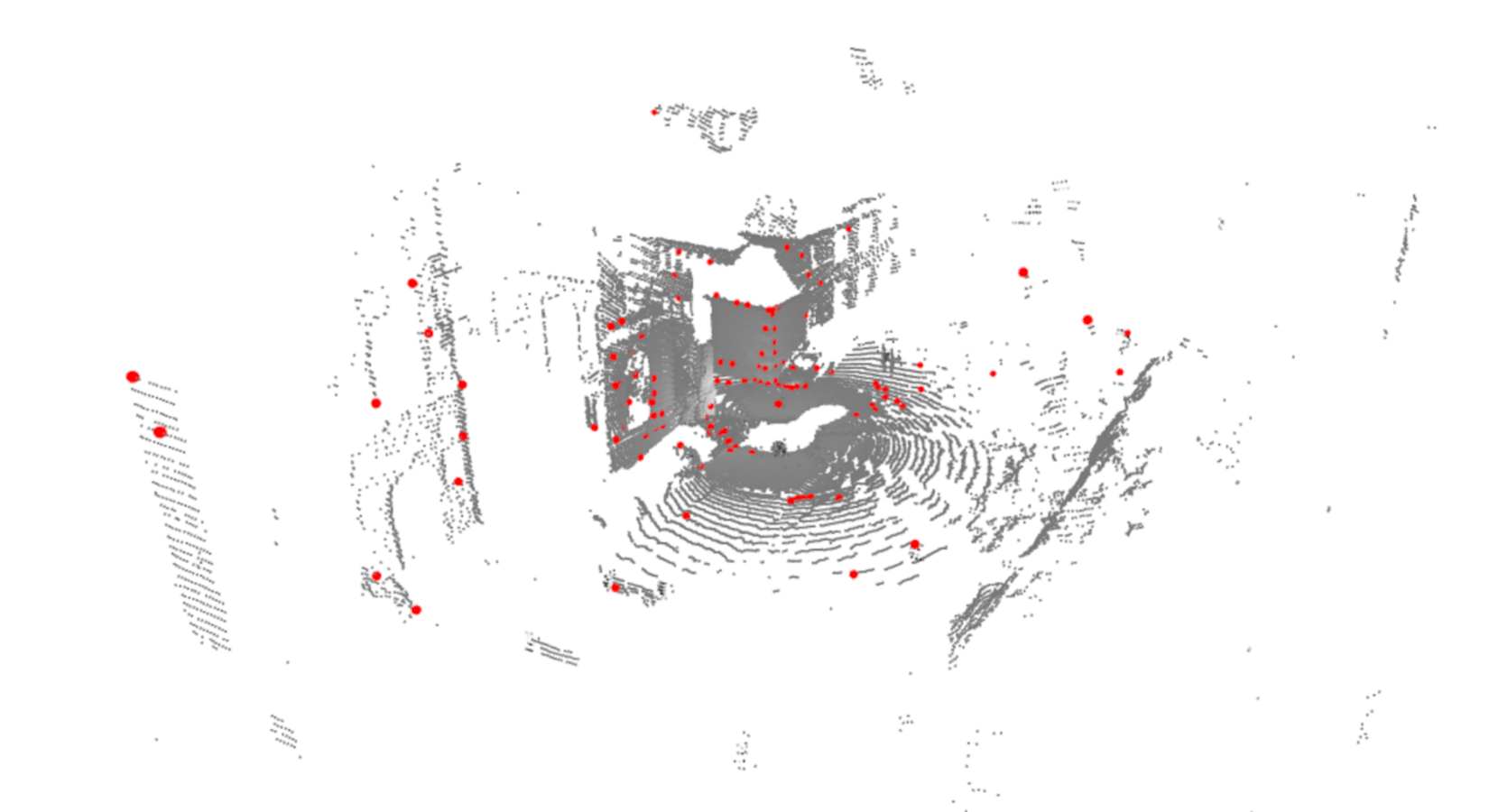}
    \end{minipage}
    \caption{Extracted keypoints (red) from an outdoor (left) and indoor (right) scene.}
    \label{fig:key_points}
    \vspace{-3mm}
\end{figure}
\begin{figure}
    \vspace{-3mm}
    \begin{minipage}{0.49\linewidth}
    \includegraphics[width=\linewidth]{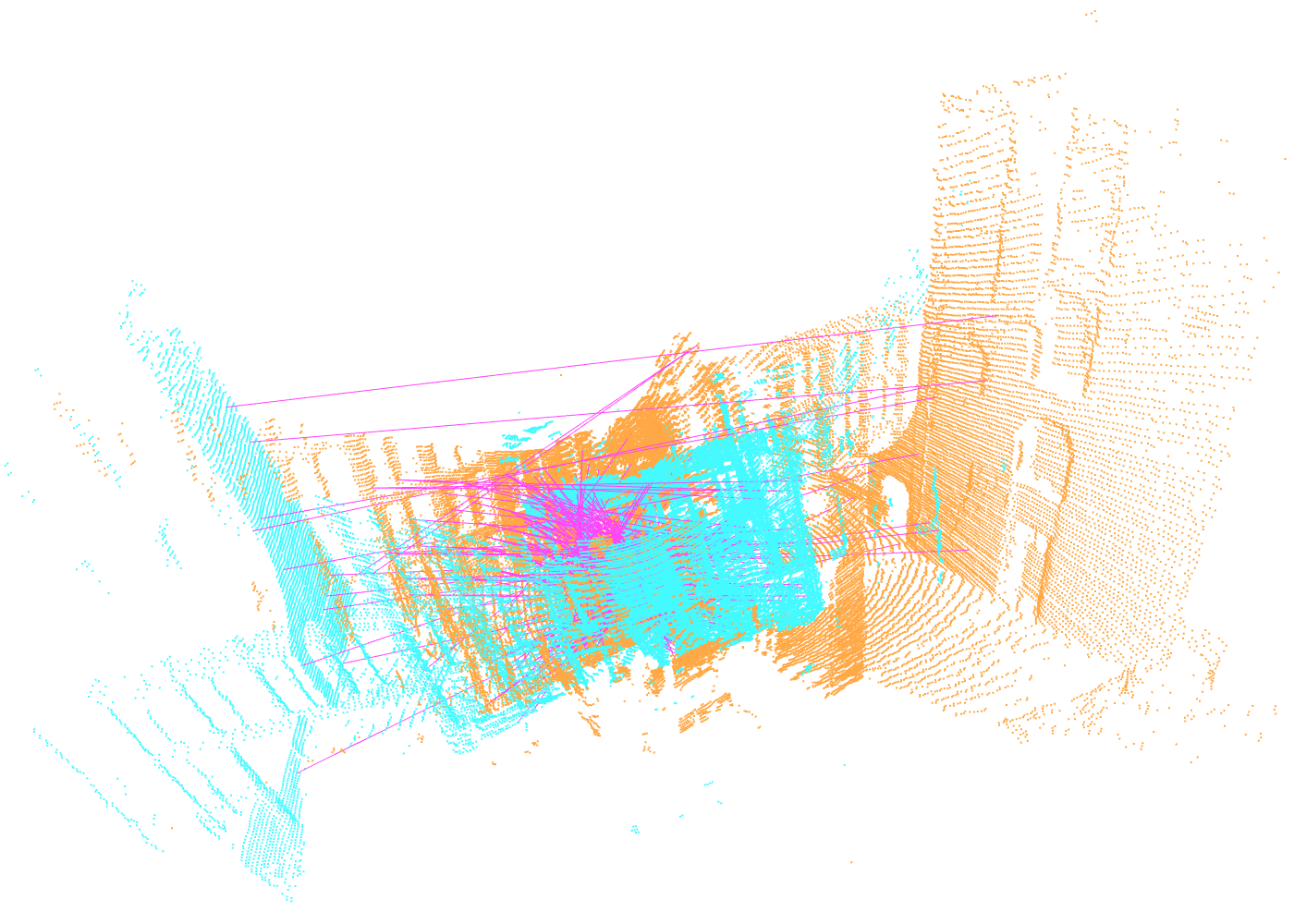}
    \end{minipage}\hfill
    \begin{minipage}{0.49\linewidth}%
    \includegraphics[width=\linewidth]{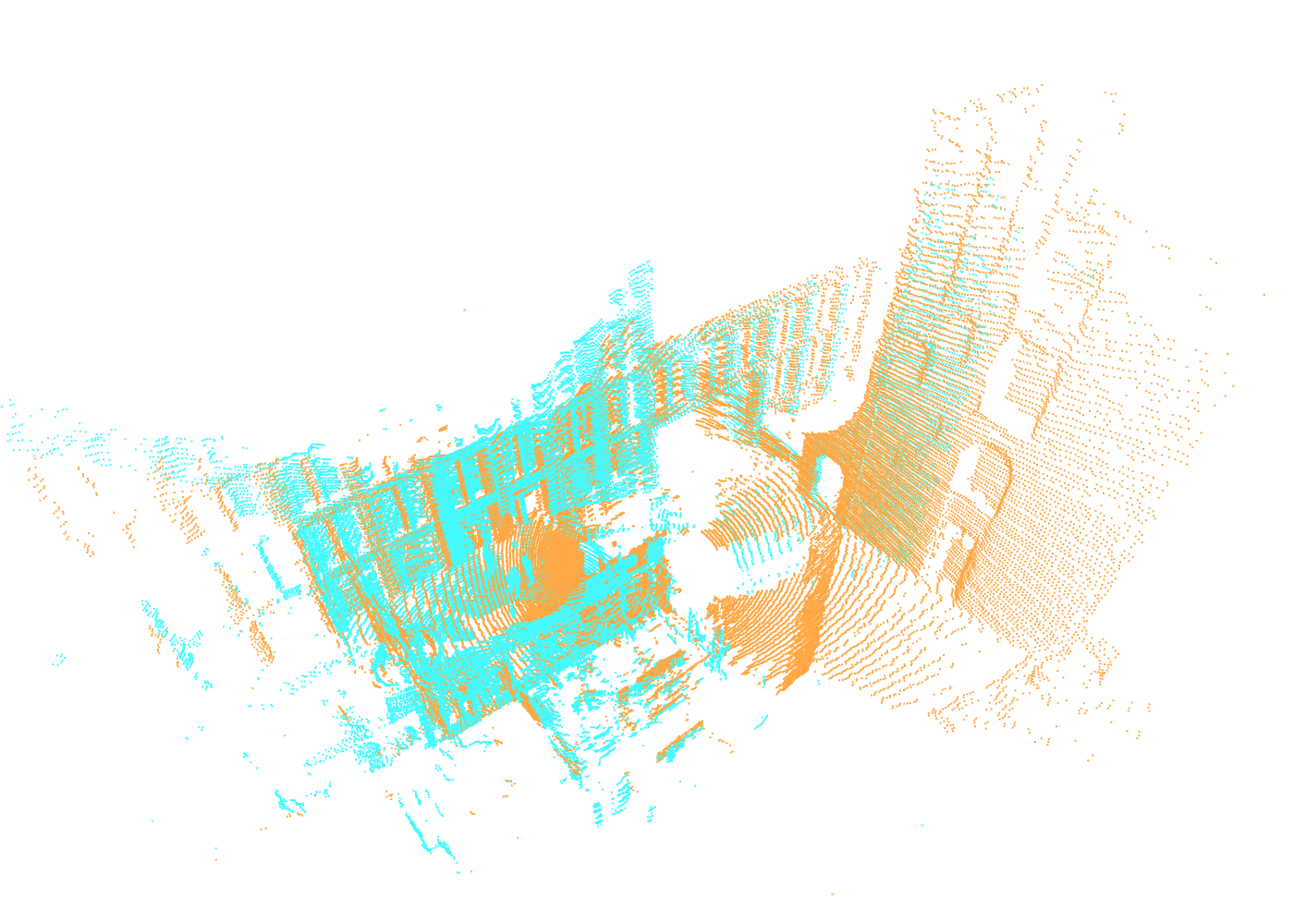}
    \end{minipage}
    \caption{Unaligned pointclouds (left) from two different LiDAR scan views (pink lines indicating found keypoint matches between them) and aligned to the same frame (right).}
    \label{fig:point_cloud_alignment}
    \vspace{-6mm}
\end{figure}
A qualitative assessment of the extracted keypoints is shown in Figure~\ref{fig:key_points} and in Figure~\ref{fig:point_cloud_alignment} an example of the pointcloud alignment is shown.
The results of the benchmark metrics are shown in Table~\ref{tab:benchmark_results}.
We compare our pipeline to the handcrafted method Harris3D~\cite{Sipiran2011} keypoints and NARF~\cite{Steder2010} descriptors, the state-of-the-art network D3Feat \cite{Bai2020} and our approach trained on synthetic pairs and real pairs.
The results show that the hand-crafted method is outperformed by all other approaches as expected. Moreover, 3DFeat~\cite{Bai2020} is also weaker than our chosen approach. However, 3DFeat~\cite{Bai2020} net is trained on the 3DMatch~\cite{Zeng2017} dataset, which includes pointclouds with uniform distribution of points in the whole scan. In contrast, in the LiDAR scan data we use, points far away are less dense than close ones. This means descriptors and the keypoint detector may not be trained for handling those pointclouds that well in the D3feat~\cite{Bai2020} case.\\
\begin{table}[H]
\vspace{-6mm}
\caption{Benchmark results of the repeatability score (RS), matching recall (MR) and registration recall (RR) in percent for keypoint extraction and description methods. dim: dimension}
\vspace{-4mm}
    \label{tab:benchmark_results}
    \begin{center}
        \begin{tabular}{lccc}
            \toprule
             & RS [\%] & MR [\%] & RR [\%]\\ 
            \midrule
            Harris3D and NARF & 23.2 & 28.3 & 4.7\\
            D3Feat & 34.6 & 29.1 & 13.4\\
            3D3L (range-only 32 dim) & 26.0 & 34.6 & 15.7\\
            3D3L (synthetic 32 dim) & 30.2 & 45.7 & 29.1\\
            3D3L (real 32 dim) & 35.4 & 60.6 & 30.7\\
            \textbf{3D3L (real 128 dim)} & \textbf{39.1} & \textbf{70.0} & \textbf{39.4}\\
            \bottomrule
        \end{tabular}
    \end{center}
    \vspace{-4mm}
\end{table}
The 3D3L network trained with range images only performs similarly to the D3-feat net, however the repeatability score is lower. This may come from missing details in the local 2D neighbourhood of the key points in the range image. The keypoint score is dependent on the distinctiveness of the descriptors (reliability) using the jointly learned approach. Therefore, this approach largely profits from the locally more detailed intensity information.
3D3L trained with synthetic pairs (32-dimensional descriptors) shows competitive results on the benchmark compared to the other approaches.
Although, the training approach with real pairs was trained on a smaller variety of data, it outperforms the synthetic pair approach.
Most likely, the model for generating synthetic pairs is not as good as training with real data.To have a comparison with what can be achieved with a larger descriptor dimension, a model with $d=128$ was trained with real pairs.
This final network outperforms all other approaches on all chosen metrics. Most of the found matches are correct, however, the keypoints are not that well localized and the registration recall is only at around $30\%$. 
Further investigations yielded a registration recall of $60.0\%$, if the threshold $\tau_3=0.5$ is increased. This shows that the results can be improved when using \ac{ICP}, since this prior is still good enough to find a precise solution. Moreover, finding a way to improve the keypoint localization may lead to even better results.\\
The runtime for extracting the features and descriptors on a Nvidia GTX 980-TI and an Intel Core I7-6700K is $\unit[22.9]{ms}$ for a single 2 channel scan image of size $\unit[64\times1024]{px}$. The time to estimate the rigid transformation with RANSAC takes $\unit[9.5]{ms}$ on average during our benchmark.
\subsection{Pose Estimation}
\label{sec:pose_estimation_results}
The evaluation of our algorithm in this section is twofold. First, we employ a scan-to-scan registration to infer an estimate for the odometry. To showcase the network's possibilities, we only use LiDAR scans at $\unit[1]{Hz}$ for odometry estimation.
 
\begin{figure}
    \vspace{2mm}
    \begin{minipage}{0.49\linewidth}
    \centering
    \includegraphics[width=\linewidth]{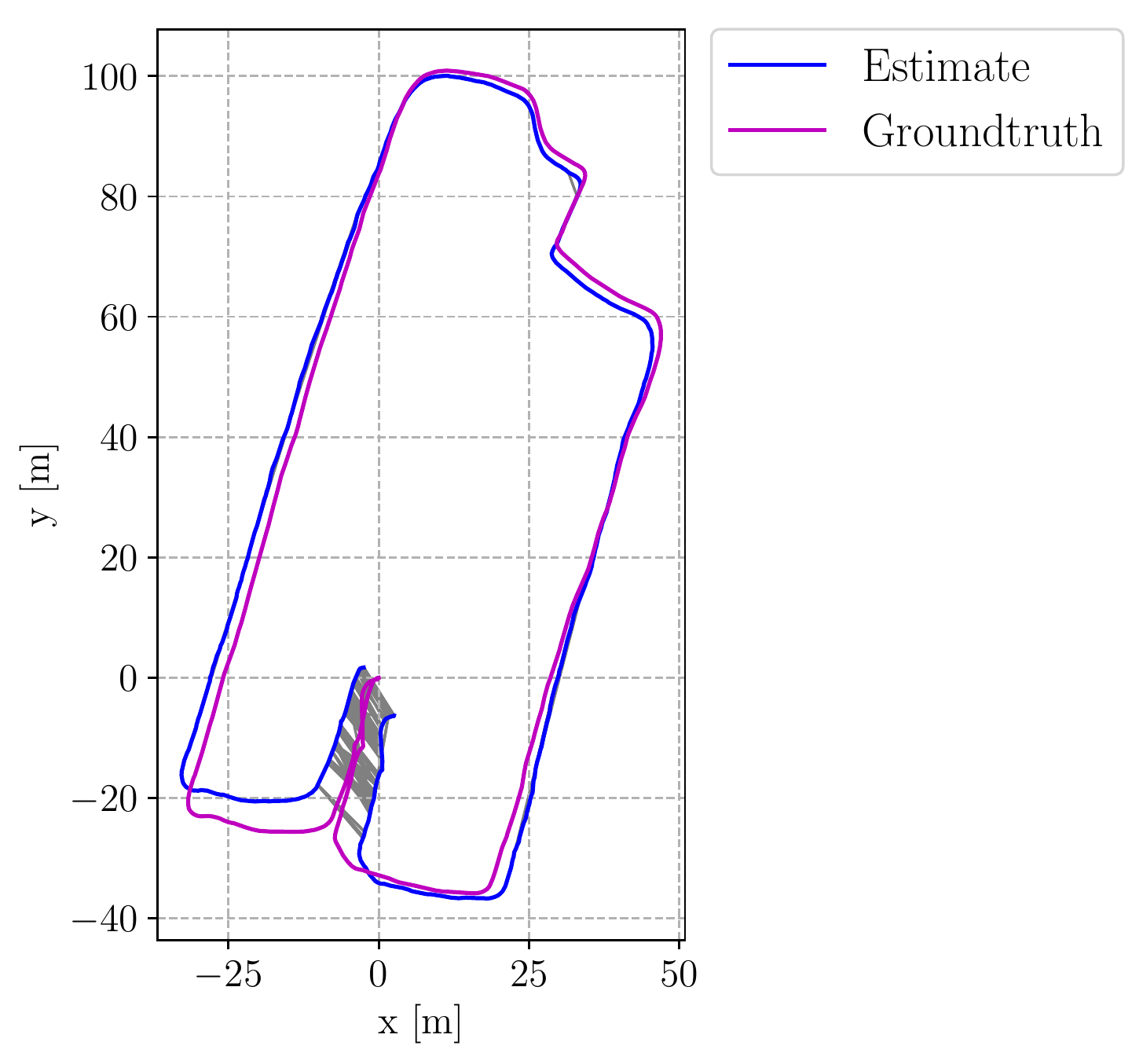}
    \end{minipage}
    \begin{minipage}{0.49\linewidth}
    \includegraphics[width=\linewidth]{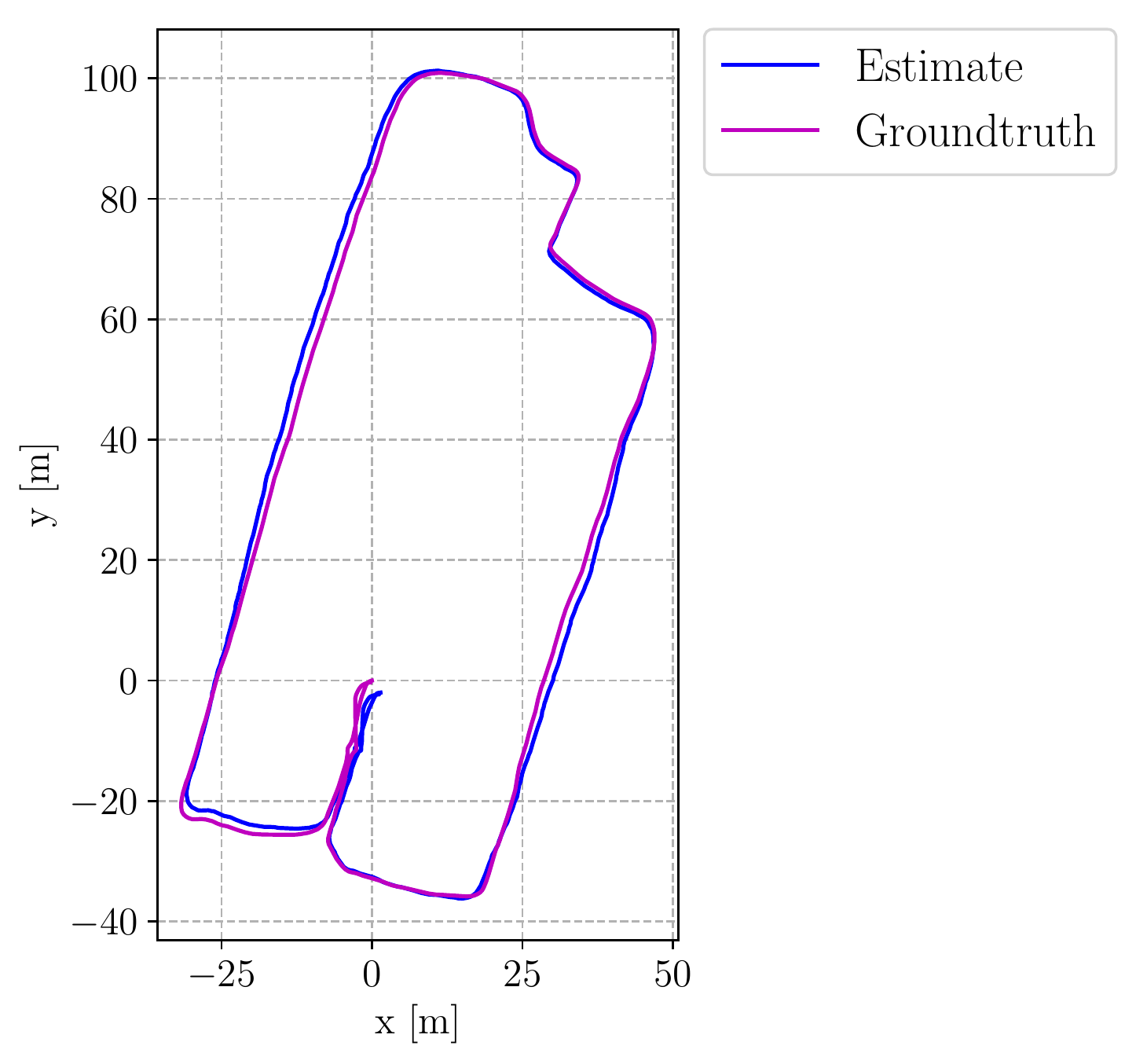}
    \end{minipage}
    \caption{Top view of trajectory from scan-to-scan pose estimation before (left) and after (right) applying loop closure (gray) edges to the optimization.}
    \label{fig:traj_top}
    \vspace{-5mm}
\end{figure}

Second, together with a \ac{BoVW}, we employ our approach as a loop closure engine. 
The results of the trajectory estimation are shown in Figure~\ref{fig:traj_top}. We have estimated the trajectory with the real pairs trained network on the ETH outside OS-0 dataset, as proposed in Section~\ref{sec:pose_estimation}. 
The following parameters are chosen: $k=180$ clusters as the word dictionary, $\tau_h=0.8$ as maximum histogram distance.
The pose estimation with only scan-to-scan estimated transformation shows a correct trend but is pretty imprecise. In Figure~\ref{fig:traj_top} (left), the trajectory shift in the $x\text{-}y$-plane is seen, where the drift of the estimation is clearly visible at the start and end of the trajectory, which should coincide.
Moreover, there is a decent amount of drift in the $z$-direction, which results most probably from the bad pitch angle estimation, resulting from a limited vertical resolution of the LiDAR scan.
Small errors in height pixel shifts lead to much larger total pose errors.
In conclusion, the mean translation error is $\unit[3.57]{m}$ and the mean rotation error is $9.21^\circ$.

Optimizing the pose graph with the additional loop closure constraints results in a substantial improvement as seen in Figure~\ref{fig:traj_top} (right).
The mean translation error is reduced to $\unit[1.23]{m}$. Moreover, the rotation error is reduced to a mean value of $5.5^\circ$.

\section{Conclusion and Future Work}
\label{sec:summary}
We presented 3D3L, a method for training an adaption of the R2D2~\cite{Revaud2019} network with LiDAR scan images, which outperforms the current state-of-the-art in keypoint extraction and description on several metrics.
We show that the combination of range and intensity channels yields robust 3D keypoints and descriptors.
Furthermore, we presented a simple pose-graph-based pipeline to show the performance of our keypoints. 
Our experiments show that it is possible to use LiDAR scan image pairs for training the network and achieve accurate pose estimation with loop closures.

For future work, we consider having a better LiDAR simulation to generate synthetic pairs that can be co-trained with real scan pairs. 
Furthermore, the range and intensity channels are inherently different, and consequently, we consider optimizing the network structure to better fit the nature of the range and intensity values.
%
%
%
%
\bibliographystyle{IEEEtran}
\bibliography{bib/references}
\end{document}